\documentclass[10pt,twocolumn,letterpaper]{article}

\usepackage{wacv}
\usepackage{times}
\usepackage{epsfig}
\usepackage{graphicx}
\usepackage{amsmath}
\usepackage{amssymb}
\usepackage{pifont}

\usepackage{multirow}
\usepackage{float}
\usepackage{subcaption}
\usepackage{ifthen}
\usepackage{xcolor}
\usepackage{verbatim}
\usepackage[accsupp]{axessibility}

\newcommand{\netname}{SpAgNet}

\def\wacvPaperID{1962} 

\wacvalgorithmstrack   

\wacvfinalcopy 

\newcommand\mr[2]{\multirow{#1}{*}{#2}}
\newcommand\nyuqt[2]{\includegraphics[width=0.15\linewidth]{imgs/nyu-qualitative/#1/#2.png}}
\newcommand\imglabel[1]{\scriptsize \textbf{#1}}

\ifwacvfinal
\usepackage[colorlinks,breaklinks=true,bookmarks=false]{hyperref}
\else
\usepackage[pagebackref=true,breaklinks=true,colorlinks,bookmarks=false]{hyperref}
\fi

\definecolor{somegray}{rgb}{0.5, 0.5, 0.5}
\newcommand{\darkgrayed}[1]{\textcolor{somegray}{#1}}
\makeatletter
\newcommand*\titleheader[1]{\gdef\@titleheader{#1}}
\AtBeginDocument{%
  \let\st@red@title\@title
  \def\@title{%
    \vskip-3em
    \bgroup\normalfont\large\centering\@titleheader\par\egroup
    \vskip1.5em\st@red@title}
}
\makeatother

\titleheader{\darkgrayed{This paper has been accepted for publication at the \\
IEEE/CVF Winter Conference on Applications of Computer Vision (WACV), Waikoloa, 2023.
\copyright IEEE}}

\title{Sparsity Agnostic Depth Completion}

\author{Andrea Conti \quad\quad\quad\quad\quad Matteo Poggi \quad\quad\quad\quad\quad Stefano Mattoccia \\%
Department of Computer Science and Engineering (DISI), University of Bologna, Italy\\
{\tt\small Project page: \url{https://andreaconti.github.io/projects/sparsity_agnostic_depth_completion} }
}

\begin{document}

\twocolumn[{
\maketitle
\vspace{-1cm}
\begin{center}
    \renewcommand{\tabcolsep}{2pt}
    \captionsetup{type=figure}
    \begin{tabular}{ccccccc}
    \rotatebox[origin=l]{90}{\scriptsize \quad \textbf{\netname{} (ours)}} & \includegraphics[width=.18\textwidth]{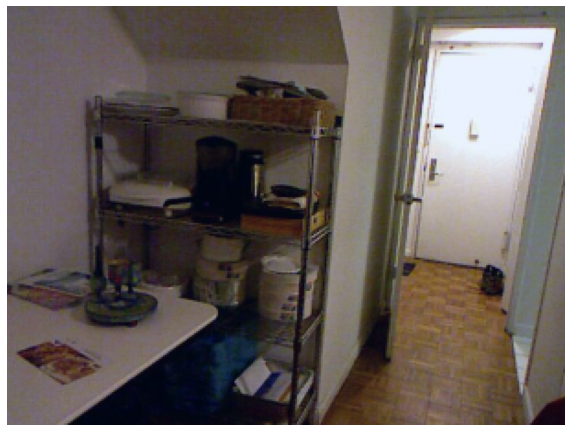} &
    \includegraphics[width=.18\textwidth]{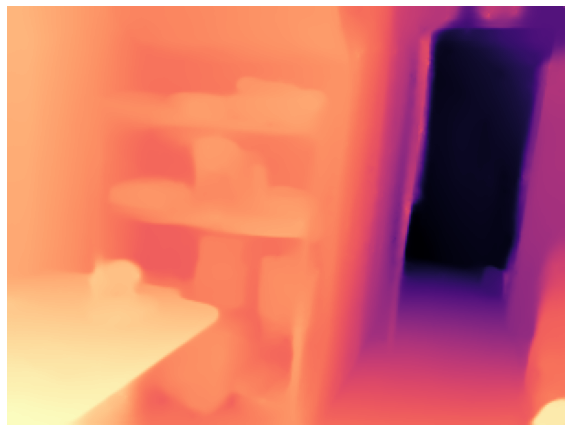} &
    \includegraphics[width=.18\textwidth]{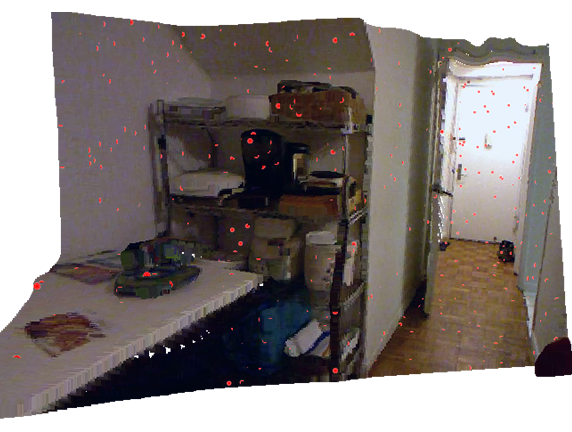} &
    \includegraphics[width=.18\textwidth]{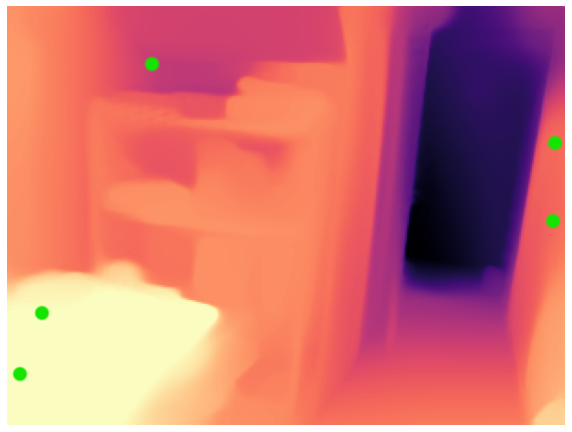} &
    \includegraphics[width=.16\textwidth]{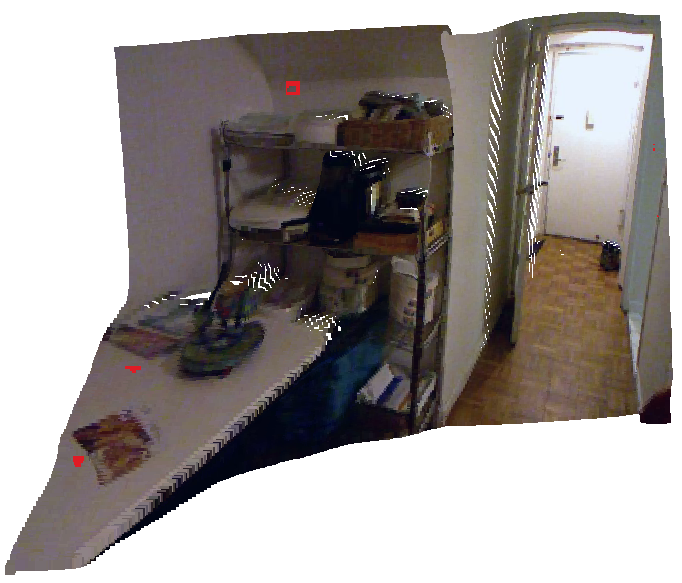} \\
    \rotatebox[origin=l]{90}{\scriptsize \quad\quad \textbf{NLSPN \cite{nlspn}}} & \includegraphics[width=.18\textwidth]{imgs/teaser/img.png} &
    \includegraphics[width=.18\textwidth]{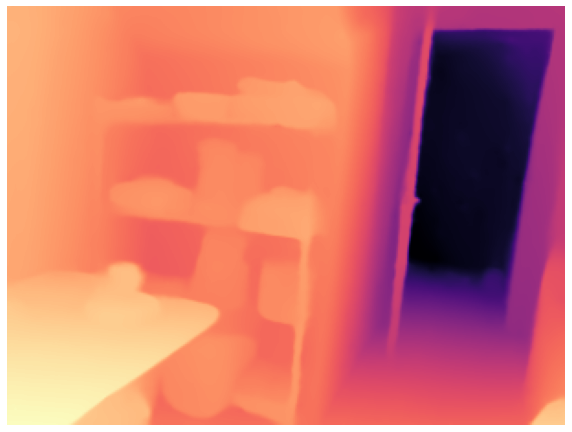} &
    \includegraphics[width=.18\textwidth]{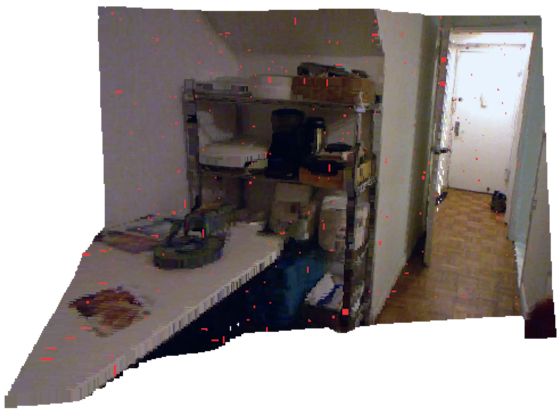} &
    \includegraphics[width=.18\textwidth]{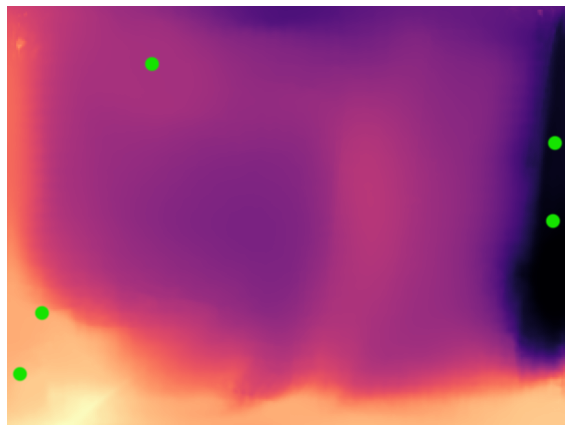} &
    \includegraphics[width=.18\textwidth]{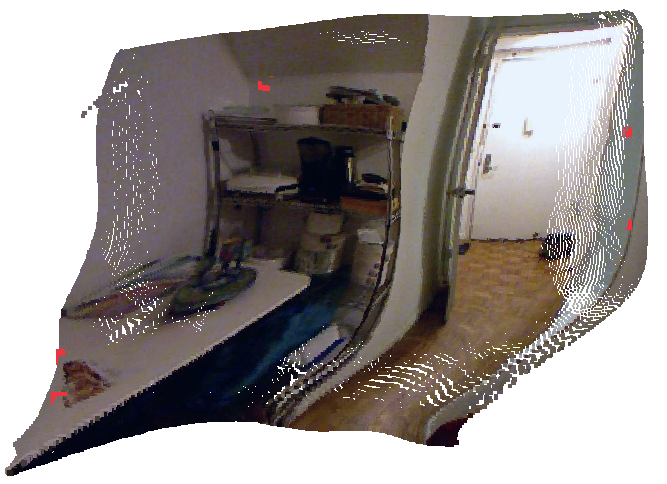} \\
    & \scriptsize \textbf{(a) RGB image} & \multicolumn{2}{c}{\scriptsize \textbf{(b) completion with 500 input points}} & \multicolumn{2}{c}{\scriptsize \textbf{(c) completion with 5 input points}} \\
    \end{tabular}
    \vspace{-0.3cm}
    \captionof{figure}{\textbf{Sparsity-agnostic depth completion.} From left to right: (a) reference image, (b) completed depth and point cloud using $500$ depth points, (c) completed depth and point cloud using only $5$ depth points. Our framework  (top) dramatically outperforms NLSPN~\cite{nlspn} (bottom) when both are trained with $500$ points and tested with much fewer.}
    \label{fig:teaser}
\end{center}
}]
\thispagestyle{empty}

\begin{abstract}
    We present a novel depth completion approach agnostic to the sparsity of depth points, that is very likely to vary in many practical applications. State-of-the-art approaches yield accurate results only when processing a specific density and distribution of input points, i.e. the one observed during training, narrowing their deployment in real use cases. On the contrary, our solution is robust to uneven distributions and extremely low densities never witnessed during training.
    Experimental results on standard indoor and outdoor benchmarks highlight the robustness of our framework, achieving accuracy comparable to state-of-the-art methods when tested with density and distribution equal to the training one while being much more accurate in the other cases. 
    Our pretrained models and further material are available in our project page. 
\end{abstract}

\section{Introduction}

Depth perception is pivotal to a variety of applications in robotics, scene understanding and more, and for this reason, it has been intensively investigated for decades. Among popular systems leveraging depth estimation, it is worth mentioning autonomous driving~\cite{Geiger2012CVPR}, path planning and augmented reality. To date, accurate depth perception is demanded either to multi-view imaging approaches \cite{mvsnet} or to specifically designed sensors such as ToF (Time of Flight) or LiDAR (Light Detection and Ranging). Although more expensive than standard cameras, depth sensors usually allow for higher accurate measurements even though at a lower spatial resolution. 
On the one hand, ToF sensors are cheap, small, and have been recently integrated into mobile consumer devices \cite{jiang_low_2022, vcsel}. They perturb the scene through coded signals unable to cope with outdoor daytime environments. To limit power consumption, a sparse emitting pattern is used, yielding meaningful depth measures for only a few points in the scene ($\sim$500 points) \cite{jiang_low_2022}. On the other hand, LiDAR sensors employ a moving array of laser emitters scanning the scene and outputting a point cloud \cite{lidarmechanism}, which becomes a sparse depth map once projected over the image camera plane due to its much higher resolution. Devices leveraging such technology are expensive and bulky however, being applicable even in daylight outdoor environments, became standard for autonomous driving applications \cite{SparsityInvariantCNN}. Since all these depth sensors provide -- for different reasons -- only sparse information, techniques aimed at recovering a dense depth map from an RGB image and a few measurements have gained much popularity in recent years \cite{SparseToDense, nlspn, cspn}.

Unfortunately, in real scenarios LiDAR and ToF sensors are affected by additional issues other than sparsity, which may easily lead even to sparser depth points often unevenly distributed. For instance, the noise originating from multi-path interference -- when multiple bouncing rays from different scene points collide on the same pixel -- might lead the sensor to invalidate the measurement and consequently reduce density. Moreover, low-reflectivity surfaces/materials absorb the whole emitted signal while others reflect it massively, leading to saturation. Despite the two opposite behaviors, depth cannot be reliably measured in both cases, possibly leading to large, unobserved regions. 

State-of-the-art depth completion techniques are fragile and fail at reconstructing the structure of the scene for areas where no depth points are available or when the sparsity changes significantly compared to the one used at training time. Indeed, the incapacity to deal with uneven spatial distributions of the sparse depth points -- which will be unveiled in this work -- threatens the possibility of deploying such solutions in different practical contexts. Moreover, this behaviour also prevents their seamless deployment when using a different sensor inferring the depth according to a spatial pattern different from the one used while training (e.g., switching from an expensive Velodyne \cite{velodyne} LiDAR system to a cheaper one).

Unfortunately, as reported in this paper and shown in Figure \ref{fig:teaser}, convolutional layers struggle at generalizing when fed with variable sparsity input data. Hence, we propose a design strategy that diverges from the literature to overcome this issue by not directly feeding sparse depth points to the convolutional layers. Purposely, we iteratively merge the sparse input points with multiple depth maps predicted by the network. 
This strategy allows us to handle highly variable data sparsity, even training the network with a constant density distribution as done by state-of-the-art methods \cite{nlspn,cspn,pncnn,packnet-san} yet avoiding catastrophic drops in accuracy witnessed by competitors. Such an achievement makes our completion solution a \textit{\underline{Sp}arsity \underline{Ag}nostic} \underline{Net}work, dubbed \netname.

Our contribution can be summarized as follows:

\begin{itemize}
    \item We propose a novel module designed to incorporate sparse data for depth completion yet being independent by their distribution and density. Such a module plugged into a competitive neural network architecture trained effortlessly can effectively deal with the previously mentioned issues.
    \item We assess the performance of \netname{} and state-of-the-art methods on a set of highly challenging cases using KITTI Depth Completion (DC) and NYU Depth V2 (NYU) datasets. We highlight the superior robustness of our solution compared to state-of-the-art when dealing with uneven input patterns.
\end{itemize}

\section{Related Work}

\textbf{Depth Prediction.} Except for a few attempts to solve monocular depth prediction through non-parametric approaches \cite{depthtransfer}, the practical ability to solve this ill-posed problem has been achieved only with the deep learning revolution. At first, deploying plain convolutional neural networks \cite{eigencnn2014} and then, through more complex approaches. Specifically, \cite{dorn} casts the problem as a classification task, \cite{banet} exploits a bidirectional attention mechanism, \cite{bts} introduces novel local planar guidance layers to better perform the decoding phase, \cite{vip-deeplab} jointly computes panoptic segmentation to improve depth prediction performance, \cite{midas} unifies multiple depth sources to coherently train a neural network to better generalize. The previous methods require a massive quantity of training data to achieve proper performance in unknown environments thus self-supervised paradigms gained much attention. For instance, \cite{monodepth2} relies on a supervisory signal extracted from a monocular video stream.

\begin{figure*}[ht] 
    \centering
    \includegraphics[width=1.00\linewidth]{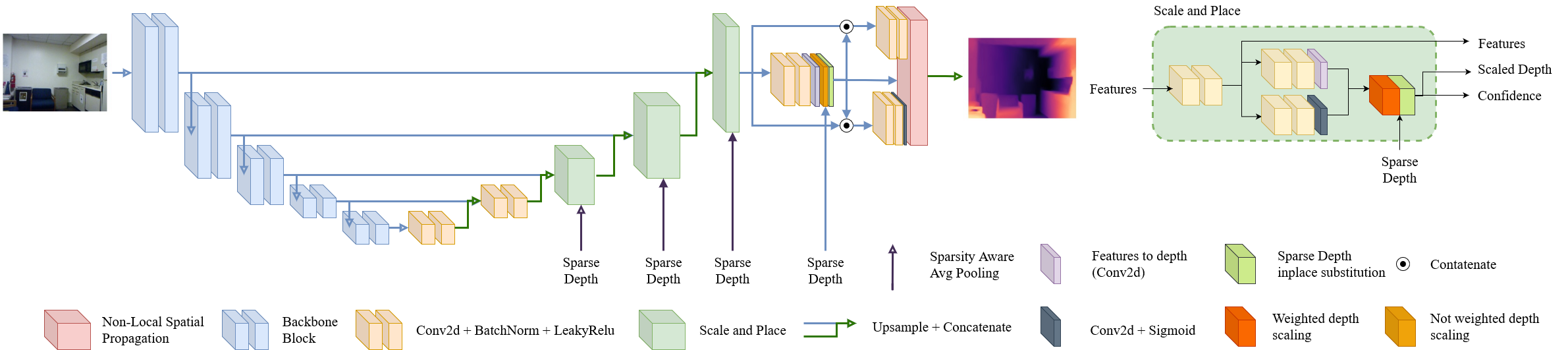}
    \caption{
        \textbf{\netname{} architecture.} The network follows an encoder-decoder design, with a backbone to extract features from the image and a custom decoder to iteratively merge at multiple scales sparse depth hints without directly feeding them as a sparse depth map. Finally, we leverage non-local propagation \cite{nlspn} to improve accuracy further.
    }
    \label{fig:network scheme}
\end{figure*}

\textbf{Depth Completion.} Depth completion aims at densifying the sparse depth map obtained by an active depth sensor, {providing sparser to denser measurements depending on the technology -- e.g., as evident by comparing Radar \cite{radardepth20} versus LiDAR \cite{Geiger2012CVPR} sensors}. This task has been tackled either by leveraging an additional RGB image or barely using the sparse depth data. Although most methods rely on learning-based paradigms, \cite{ZhaoYiming_IEEE_2021} proposes a non-parametric handcrafted method.   
Regarding deep-learning methods, \cite{SparseToDense} was among the first to tackle the problem by jointly feeding a neural network with the RGB frame and the sparse depth points to densify the latter. Observing that manipulating sparse data is sub-optimal for convolutions, \cite{SparsityInvariantCNN, MinkoswkiConvs} proposed custom convolutional layers explicitly taking into account sparsity. 
Eventually, guided spatial propagation techniques have demonstrated superior performance. At first, \cite{spn} proposed a network able to learn local affinities to guide the depth expansion, this strategy was improved initially by \cite{cspn} and then by \cite{nlspn}.
Based on a similar principle, \cite{GuideNet} proposes content-dependent and spatially-variant kernels for multi-modal feature fusion.  \cite{pncnn} performs depth completion also modeling the confidence of the sparse input depth and the densified output. 
{In a parallel track, a few works focused on unsupervised training strategies for depth completion \cite{ma2018self,wong2020unsupervised,wong2021learning,wong2021unsupervised}.}
Finally, \cite{packnet-san} proposes an approach to deal with depth completion and depth prediction. Even though this seems similar to our research, it is only loosely related. First, it cannot deal with different sparsities but only with the total absence of the sparse depth points. Second, to achieve their goal, they need a specific training procedure and an additional branch to handle the availability of sparse depth data. In contrast, our peculiar network design addresses both issues.

\textbf{Uncertainty Estimation.} Evaluating the estimated value's uncertainty (or confidence) is essential in many circumstances. For neural networks, it has been widely explored either the use of Bayesian frameworks \cite{BayesianFramework, BayesianLearning, pmlr-v32-cheni14} or strategies jointly predicting the mean and variance of the network's output distribution \cite{PredictConfidence}. For depth completion, \cite{pncnn} proposed to jointly compute the confidence of the sparse input depth and of the densified output. While, for monocular depth prediction,  \cite{SelfSupervisedConfidence} has deeply investigated uncertainty for self-supervised approaches.

\section{Sparsity Agnostic Framework}
\label{sec:our method}

To tackle depth completion, we start from our previous observations. Specifically, as pointed out by \cite{SparsityInvariantCNN, MinkoswkiConvs}, 2D convolutions struggle to manipulate sparse information. Additionally, we further notice that the density of such input depth data and its spatial distribution -- which could be highly uneven -- might lead state-of-the-art networks to catastrophic failures, as depicted at the bottom of Figure \ref{fig:teaser}. Moreover, we argue that these networks mostly rely on the sparse depth input overlooking the image content substantially ignoring the geometric structure depicted in it.    

\netname{} relies on an encoder-decoder structure with skip connections, as depicted in Figure \ref{fig:network scheme}. However, unlike current depth completion techniques \cite{nlspn, cspn, packnet-san, pncnn}, we do not feed the encoder with sparse depth information for the reasons previously outlined. We extract instead features from the RGB frame \textit{only} in order to get rid of the sparse input data and, consequently, its density. This strategy allows us to constrain the network to exploit the image content fully and, as we will discuss later, to enforces the network to extract the geometry of the scene from RGB. 

The decoding step predicts -- iteratively and at multiple scales -- dense depth from the RGB image and \textit{fuse} it with the sparse input data. The first iterative step takes the input features extracted from the RGB image and generates a lower scale depth map and a confidence map. Then, the next iterative steps process the same inputs plus the depth map and its confidence, both \textit{augmented} with the sparse input points computed in the previous iteration. Moreover, since each intermediate depth map provides information up to a scale factor, we scale it according to the sparse input points before each augmenting step. We do so due to the ill-posed nature of monocular depth prediction. Experimental results will corroborate our design choice, especially when dealing with a few sparse input points.
At the end of the iterative steps, we apply the non-local spatial propagation module proposed in \cite{nlspn} to refine the depth map inferred by the network. Figure \ref{fig:network scheme} describes the whole framework.

\subsection{Encoder Architecture}
\label{subsec:encoder}

Since our framework encodes features from the image only, we can leverage as encoding backbone any pre-trained network. Such backbone is pre-trained on ImageNet \cite{ImageNet}. Among the multiple choices \cite{resnet, densenet, ResNeXt50} we choose ResNeXt50 \cite{ResNeXt50} due to its good trade-off between performance and speed. Specifically, it downsamples the image to scales $\frac{1}{2}$, $\frac{1}{4}$, $\frac{1}{8}$, $\frac{1}{16}$ and $\frac{1}{32}$ and the features used in the decoding step as input and skip connection.

\begin{figure*}[t]
    \centering
    \renewcommand{\tabcolsep}{2pt}
    \begin{tabular}{ccc}
        \begin{tabular}{cc}
            \includegraphics[width=0.15\linewidth]{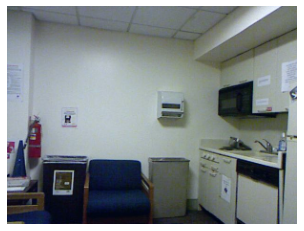} & 
            \includegraphics[width=0.15\linewidth]{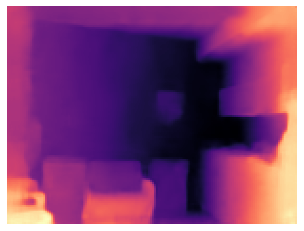} \\
            \scriptsize \textbf{(a) RGB} & \scriptsize \textbf{(b) dense depth} \\
            \includegraphics[width=0.15\linewidth]{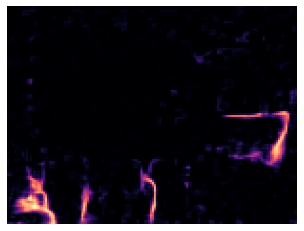} &
            \includegraphics[width=0.15\linewidth]{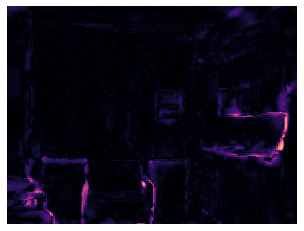} \\
            \scriptsize \textbf{(c) confidence map} & \scriptsize \textbf{(d) depth error} \\
        \end{tabular} & \quad\quad & 
        \begin{tabular}{c}
            \includegraphics[width=0.6\linewidth]{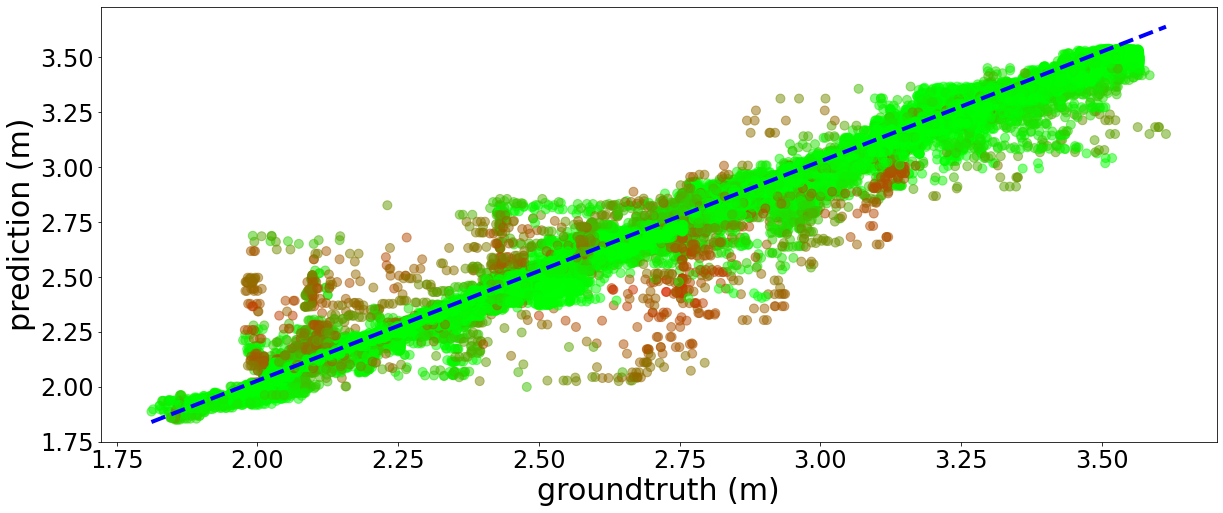} \\
            \scriptsize \textbf{(e) depth scaling linear regression} \\
        \end{tabular}  
    \end{tabular}
    
    \caption{
    \textbf{Confidence aware depth scaling.} Example of confidence usage to scale depth. On left, we show (a) the input image, (b) predicted depth map, (c) estimated confidence and (d) errors with respect to groundtruth. On right, we plot the outcome of the scaling procedure (red means a lower confidence prediction, green a higher one). 
    }
    \label{fig:confidence}
\end{figure*}

\subsection{Scale and Place Module}
\label{subsec:scale and position}

In our proposal, the core Scale and Place (S\&P) module is in charge of inferring a dense and scaled depth map and its confidence. It takes as input the backbone features, the output of the previous S\&P module at a different scale, and the sparse depth points as depicted in Figure \ref{fig:network scheme}. 

Specifically, S\&P leverages the input features to jointly generate an initial up-to-scale depth map and its confidence deploying a stem block composed of two convolutional layers and two heads in charge of generating them. Each convolutional layer consists of a 2D convolution, a batch normalization \cite{batchnorm} and a Leaky ReLU. Then in the \textit{Scale} step, the S\&P module performs a weighted linear regression to scale the depth map according to the available sparse input points, weighted by means of confidence. The parameters of the weighted linear regression can be computed in closed form and in a differentiable way, as described in Eq. \ref{eq:weighted linear regression} where $p_i$ is the predicted depth value and $c_i$ its confidence corresponding to an available input sparse point $s_i$.

\begin{align}
    & \beta  = \frac{\sum_ic_i(p_i-\hat p)(s_i - \hat s)}{\sum_ic_i(p_i - \hat p)^2} \quad \alpha = \hat s - \beta \hat p 
    \label{eq:weighted linear regression} \\
    & \hat p = \frac{\sum_ic_ip_i}{\sum_ic_i} \nonumber \quad \hat s = \frac{\sum_ic_is_i}{\sum_ic_i} \nonumber
\end{align}

Then, in the \textit{Place} step, for those points where a sparse input depth value is available, we replace the corresponding value in the scaled depth map with it. Additionally, we update the same point in the confidence map with the highest score. The latter step can be summarized as follows

\begin{equation}
    \hat D[x, y] = \left\{ \begin{aligned} 
      & D^s[x, y]   \quad \operatorname{if} \quad H[x, y] = 0   \\
      & H[x, y]   \ \quad \operatorname{if} \quad H[x, y] \ne 0 
    \end{aligned} \right.
    \label{eq:place step}
\end{equation}

\begin{equation}
    \hat C[x, y] = \left\{ \begin{aligned} 
      & C^s[x, y]   \quad \operatorname{if} \quad H[x, y] = 0   \\
      & 1   \quad \quad \ \quad \operatorname{if} \quad H[x, y] \ne 0 
    \end{aligned} \right.
    \label{eq:place conf step}
\end{equation}

where $D^s$ is the scaled depth map, $C^s$ is the confidence map and $H$ is a sparse depth map containing zeros where an input sparse depth point is not available. The predicted confidence has an empirically chosen range of [0.1 .. 0.9] while we associate confidence 1 to each valid value in $H$.

We apply the S\&P module at scales $\frac{1}{8}$, $\frac{1}{4}$ and $\frac{1}{2}$. The module at $\frac{1}{8}$ computes the initial depth and confidence maps leveraging only the RGB features. The others take in input also the up-sampled dense depth and confidence maps from the previous module in order to iteratively correct the prediction relying on both the predicted depth and the injected sparse points. Thus, with this strategy, the decoder does not deal directly with sparse data in any of its steps. Nonetheless, the network can locate and effectively leverage reliable sparse information.
An example of this mechanism is showed in Figure \ref{fig:confidence}, where can be clearly seen how the network learns to locate the most reliable depth values as those closer to the groundtruth depth.

It is worth noting that confidence plays a crucial role in the S\&P module. At first, in the \textit{Scale} step, it helps to locate outliers in the estimated depth map enabling to soften their impact when performing the scaling procedure. Additionally, in the \textit{Place} step, assigning the highest confidence to the sparse input points enables the network to rely on them effectively. Nonetheless, \netname{} also exploits the other predicted depth points according to their estimated confidence.

Since the S\&P module needs the sparse data at multiple scales, we down-sample it by employing a non-parametric sparsity aware pooling: moving a 3$\times$3 window with stride 2, we assign the mean of the available measures in its neighbourhood to each coordinate, we iteratively apply this process to reach lower resolutions. This approach leads to a densification of the sparse depth map and helps, at all scales, to include even the meagre few sparse points available to a large field of view.

\subsection{Non-Local Spatial Propagation}
\label{subsec:nlspn}

Spatial propagation concerns the diffusion of information in a localized position to its neighbourhoods. This strategy represents a common practice in the depth completion literature \cite{spn, cspn, nlspn, penet} and can be achieved by a neural network in charge of learning the affinity among neighbours. Let $X~=~(x_{m,n})~\in~R^{M \times N}$ be a 2D depth map to be refined through propagation, at step $t$ it acts as follows:

\begin{equation}
    x^t_{m,n} = w_{m,n}^cx^{t-1}_{m,n} + \sum_{(i,j)\in N_{m,n}}w^{i,j}_{m,n}x_{i,j}^{t-1}
\end{equation}

Where $(m, n)$ is the reference pixel currently being updated, $(i, j) \in N_{m, n}$ the coordinate of the pixels in its neighborhood, $w^{i,j}_{m,n}$ the affinity weights, and $w^c_{m, n}$ the affinity weight of the reference pixel:

\begin{equation}
    w_{m,n}^c = 1 - \sum_{(i,j)\in N_{m,n}} w^{i,j}_{m,n}
\end{equation}

The various existing methods differ by the choice of the neighborhood and by the normalization procedure of the affinity weights, the latter necessary to ensure stability during propagation \cite{spn, cspn, nlspn}. Within \netname, we implement the non-local approach \cite{nlspn}, letting the network dynamically decide the neighborhood using deformable convolutions \cite{DeformConv}. Formally:

\begin{gather}
    N_{m,n} = \{x_{m+p,n+q} \ | \ (p, q) \in f_\phi(\operatorname{I}, \operatorname{H}, n, m)\} \\
    p, q \in \mathbb{R} \nonumber
\end{gather}

Where I and H are the RGB image and the sparse depth, and $f_\phi(\cdot)$ is the neural network determining the neighbourhood. 
The non-local propagation module requires in input an initial depth map generated through two convolutional blocks from the last S\&P block output, scaled using the full-resolution sparse depth points. However, in this case, we do not perform a weighted scaling to obtain the best result on the entire frame. Finally, as usual, the sparse depth points override the predicted output. The resulting depth map is then fed along with features to two convolutional blocks to generate the guiding features and confidence required by the propagation module.

\subsection{Loss Function}

At each scale, we train the network by supervising the depth obtained by the S\&P module before \textit{Place} step. The confidence weights the loss of each depth prediction, and a regularization term (controlled by $\eta$) enforces the network to maintain the confidence as higher as possible. Following \cite{nlspn}, we compute both L1 and L2 losses. Our loss function, at a specific scale, is described by Eq. \ref{eq:loss function} where $C^s$ and $D^s$ are respectively confidence and depth at a specific scale $s$. Confidence is not computed for the full size scale, hence $C^0 = 1$. Finally, it is worth mentioning that lower scales are weighted less through an exponential decay factor $\gamma$.

\begin{align}
    & L = \sum_{s=0}^n \gamma^s \frac{1}{N}\sum_i^mC^s_iL_i^{12} - \eta\ln{C^s_i} \label{eq:loss function} \\
    & L_i^{12} = |D_i^s - G_i| + |D_i^s - G_i|^2 \nonumber
\end{align}

\section{Experimental Results}
\label{sec:experimental results}

We have implemented \netname{} in PyTorch \cite{pytorch} training with 2 NVIDIA RTX 3090 and using the ADAM optimizer \cite{adam} with $\beta_1 = 0.9$ and $\beta_2 = 0.999$. The final model requires 35 milliseconds to perform a prediction on a image of 640$\times$480 resolution employing a single NVIDIA RTX 3090 GPU.

\subsection{Datasets}

\textbf{NYU Depth V2.} The NYU Depth V2 \cite{nyu-depth} dataset is an indoor dataset containing 464 indoor scenes gathered with a Kinect sensor. We follow the official train/test split as previous works relying on the pre-processed subset by Ma et al. \cite{SparseToDense} using 249 scenes for training ($\sim$50K samples) and 215 scenes (654 samples) for testing. Each image has been down-sampled to 320$\times$240 and then center cropped to 304$\times$228. As a common practice on this dataset, 500 random points per image have been extracted to simulate sparse depth. We train our network for 15 epochs starting with a learning rate $10^{-3}$ and decreasing it every 3 epochs by $0.1$, setting $\gamma = 0.4$ and $\eta = 0.1$. We use batch size 24 (12 for each GPU); hence the network is extremely fast to converge since the whole training accounts less than 30K steps. We apply color and brightness jittering and horizontal flips to limit overfitting.

\textbf{KITTI Depth Completion (DC)}. KITTI DC \cite{SparsityInvariantCNN} is an outdoor dataset containing over 90K samples, each one providing RGB information and aligned sparse depth information (with a density of about 5\%) retrieved by a high-end Velodyne HDL-64E LiDAR sensor. The images have 1216$\times$352 resolution, and the dataset provides a standard split to train (86K samples), validate (7K samples) and test (1K samples). The groundtruth has been obtained temporally accumulating multiple LiDAR frames and filtering errors \cite{SparsityInvariantCNN}, leading to a final density of about 20\%. On this dataset we train for 10 epochs with batch size 8 (4 for each GPU), starting with learning rate $10^{-3}$ and we decrease it every 3 epochs by $0.1$, we set $\gamma = 0.4$ and $\eta = 20.0$. Data augmentation follows the same scheme used for NYU.

\subsection{Evaluation}

In this section, we assess the performance of our proposal and state-of-the-art methods deploying the dataset mentioned above. Following standard practice \cite{nlspn, cspn}, we use the following metrics: $\operatorname{RMSE}~=~\sqrt{\frac{1}{N}\sum_i|D_i - G_i|^2}$, $\operatorname{MAE}~=~\frac{1}{N}\sum_i|D_i - G_i| $ and $\operatorname{REL}~=~\frac{1}{N}\sum_i\left| \frac{D_i - G_i}{G_i}\right|$.

\begin{figure}
    \centering
    \begin{tabular}{@{}c@{}c@{}c@{}c}
    \begin{subfigure}[b]{0.25\linewidth}
        \includegraphics[width=\linewidth]{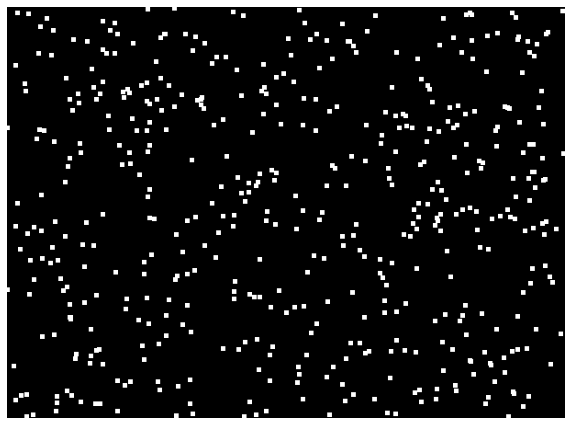}
    \end{subfigure} &
    \begin{subfigure}[b]{0.25\linewidth}
        \includegraphics[width=\linewidth]{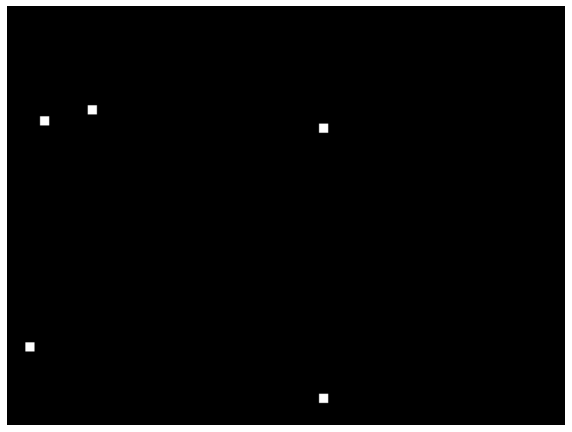}
    \end{subfigure} &
    \begin{subfigure}[b]{0.25\linewidth}
        \includegraphics[width=\linewidth]{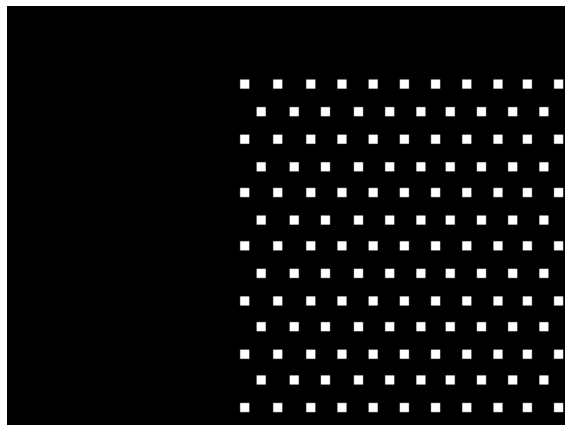}
    \end{subfigure} &
    \begin{subfigure}[b]{0.25\linewidth}
        \includegraphics[width=\linewidth]{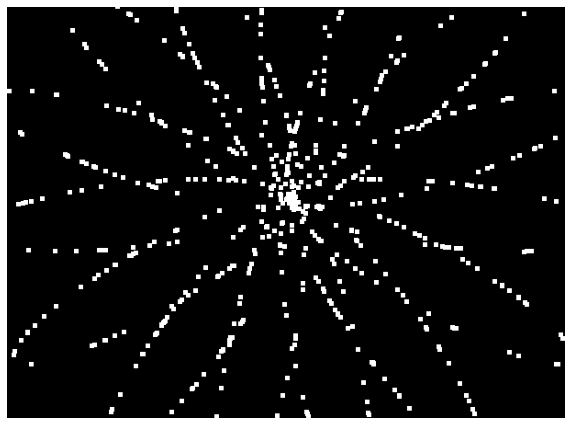}
    \end{subfigure} \\
    \imglabel{500p} & \imglabel{5p} & \imglabel{shifted grid} & \imglabel{Livox} \\
    \end{tabular}
    \caption{\textbf{Sparse depth patterns.} Examples of different sparse depth patterns, from left to right: 500 random points, 5 random points, shifted triangular tiling dot pattern and a Livox-like pattern (e.g. Livox Mid-70).}
    \label{fig:nyu-patterns}
\end{figure}

For evaluation purposes, in addition to the standard protocol deployed in this field \cite{nlspn, cspn}, we also thoroughly evaluate the robustness of the networks on the two datasets in much more challenging scenarios but always training with the standard procedure (i.e., using 500 points on NYU and 64 LiDAR lines on KITTI). 
Since KITTI DC is thought for autonomous driving tasks and the sparse depth is acquired with an high end 64 Lines Lidar which provides in output always the same pattern, we simulate the switch to a cheaper device providing in output less lines assessing the capability of \netname{} to generalize over sparse depth density. 
On NYU Depth V2, sparse depth points are traditionally extracted randomly from the groundtruth \cite{SparseToDense, nlspn, cspn} which is almost dense. Thus, we test i) the extreme case of having only 5 random points, ii) the impact of having large empty areas and iii) the impact of changing the sparsity pattern. We implement ii) sampling from the groundtruth a triangular tiling dot pattern aimed at simulating the output of a commercial VCSEL \cite{vcsel} ToF sensor and then randomly shifting this pattern to leave behind large empty areas where no sparse hints are available while iii) extracting from the groundtruth sparse points with the pattern of a Livox Mid-70 \cite{livox}. All these patterns are showed in Figure \ref{fig:nyu-patterns}. We take into account the publicly pre-trained state-of-the-art models available either on NYU Depth V2 or KITTI DC and we take care to guarantee that each architecture sees exactly the \textit{same} sparse points while being evaluated.

\begin{table}[t]
    \centering
    \scalebox{0.70}{
    \begin{tabular}{l | c | c c}
        \hline
        Method & Samples & REL $\downarrow$  & RMSE (m) $\downarrow$ \\
        \hline
        pNCNN \cite{pncnn}               & \mr{5}{500}  & 0.026             & 0.170             \\
        CSPN \cite{cspn}                 &              & 0.016             & 0.118             \\
        NLSPN \cite{nlspn}               &              & \textbf{0.013}    & \textbf{0.101}    \\
        PackNet-SAN \cite{packnet-san}   &              & 0.019             & 0.120             \\
        \netname{} (ours)                            &              & \underline{0.015} & \underline{0.114} \\
        \hline      
        pNCNN \cite{pncnn}               & \mr{5}{200}  & 0.040             & 0.237             \\
        CSPN \cite{cspn}                 &              & 0.027             & 0.177             \\
        NLSPN \cite{nlspn}               &              & \textbf{0.019}    & \textbf{0.142 }   \\
        PackNet-SAN \cite{packnet-san}   &              & 0.027             & \underline{0.155} \\
        \netname{} (ours)                            &              & \underline{0.024} & \underline{0.155} \\
        \hline      
        pNCNN \cite{pncnn}               & \mr{4}{100}  & \underline{0.061} & 0.338             \\
        CSPN \cite{cspn}                 &              & 0.067             & 0.388             \\
        NLSPN \cite{nlspn}               &              & \textbf{0.038}    & \underline{0.246} \\
        \netname{} (ours)                            &              & \textbf{0.038}    & \textbf{0.209}    \\
        \hline      
        pNCNN \cite{pncnn}               & \mr{4}{50}  & 0.108             & 0.568             \\
        CSPN \cite{cspn}                 &             & 0.185             & 0.884             \\
        NLSPN \cite{nlspn}               &             & \underline{0.081} & \underline{0.423} \\
        \netname{} (ours)                            &             & \textbf{0.058}    & \textbf{0.272}    \\
        \hline                  
        pNCNN \cite{pncnn}               & \mr{4}{5}  & 0.722             & 2.412              \\
        CSPN \cite{cspn}                 &            & 0.581             & 2.063              \\
        NLSPN \cite{nlspn}               &            & \underline{0.262} & \underline{1.033}  \\
        \netname{} (ours)                            &            & \textbf{0.131}    & \textbf{0.467}     \\
        \hline       
        \multicolumn{4}{c}{ }\\
        \hline
        pNCNN \cite{pncnn}               &                        & 0.519             & 1.922               \\
        CSPN \cite{cspn}                 & \mr{1}{shifted grid}   & 0.367             & 1.547               \\
        NLSPN\cite{nlspn}                & \mr{1}{($\sim100$)}      & \underline{0.175} & \underline{0.796} \\
        \netname{} (ours)                            &                        & \textbf{0.110}    & \textbf{0.422}      \\
        \hline
        pNCNN \cite{pncnn}               &                        & 0.061             & 0.333             \\
        CSPN \cite{cspn}                 & \mr{1}{livox}          & 0.066             & 0.376             \\ 
        NLSPN \cite{nlspn}               & \mr{1}{($\sim150$)}    & \textbf{0.037}    & \underline{0.233} \\
        \netname{} (ours)                            &                        & \underline{0.039} & \textbf{0.206}    \\
        \hline
    \end{tabular}
    }
    \caption{\textbf{Evaluation on NYU Depth v2.} Comparison with state-of-the-art methods, trained with 500 random points, extracted from groundtruth, as input and tested with different densities and patterns. In bold is the best result, underlined the second one.}
    \label{tab:sota comparison nyu}
\end{table}

\textbf{Results on NYU Depth v2.} Table \ref{tab:sota comparison nyu} compares state-of-art methods and our proposal on the NYU dataset using different input configurations: in the upper portion by changing the number of samples and in the lower portion by changing the pattern type.
From the table, we can notice that our proposal achieves competitive results, being very close to NLSPN and better than other methods when the number of points used is the same as the training phase (i.e., 500). Similar behaviour occurs with 200 points. However, when the density of input points decreases further, \netname{} vastly outperforms the state-of-the-art. The performance gap with other methods gets much higher when decreasing the density further. 
For instance, with 50 points, the RMSE by \netname{} is 0.272 m, while the second one (NLSPN) accounts for 0.423 m. Notably, with only 5 points, the same metrics are 0.467 m and 1.033 m (NLSPN), further emphasizing the ability of our proposal to deal even with meagre input points, in contrast to our competitors. It is worth observing that our method outperforms competitors with randomly selected input points starting from 100. 

\begingroup
\begin{figure*}[ht!]
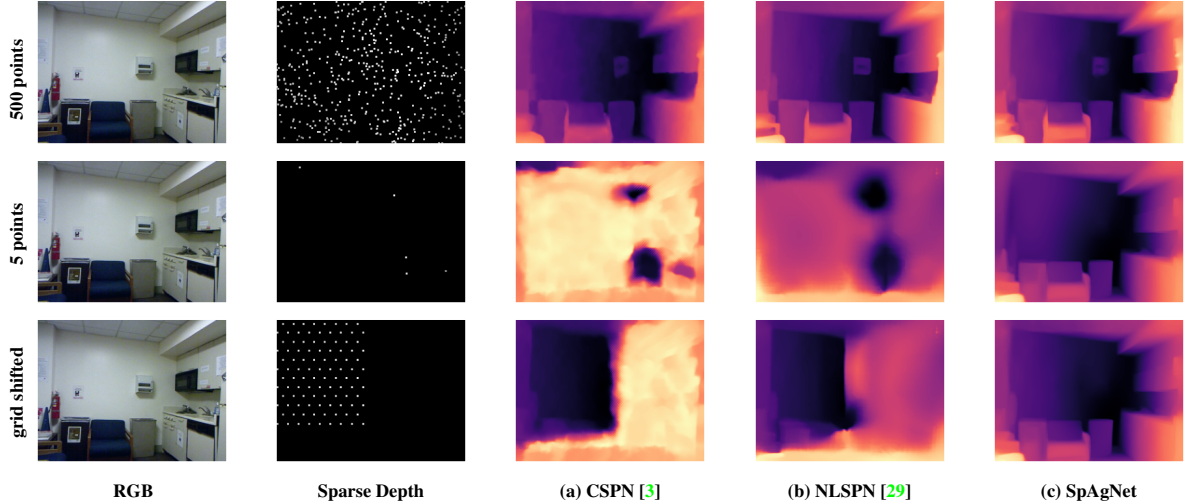

    \centering
    \renewcommand{\tabcolsep}{8pt}
    \begin{tabular}{cccccc}
        \rotatebox[origin=l]{90}{\scriptsize \quad\ \textbf{500 points}} 
        \nyuqt{our}{rgb} & \nyuqt{hints}{500p}     & \nyuqt{cspn}{500p} & \nyuqt{nlspn}{500p} & \nyuqt{our}{500p} \\
        \rotatebox[origin=l]{90}{\scriptsize \quad\quad \textbf{5 points}}  
        \nyuqt{our}{rgb} & \nyuqt{hints}{5p}       & \nyuqt{cspn}{5p}   & \nyuqt{nlspn}{5p}   & \nyuqt{our}{5p}   \\
        \rotatebox[origin=l]{90}{\scriptsize \quad\ \textbf{grid shifted}} 
        \nyuqt{our}{rgb} & \nyuqt{hints}{grid}     & \nyuqt{cspn}{grid} & \nyuqt{nlspn}{grid} & \nyuqt{our}{grid} \\
        \quad \imglabel{RGB}   & \imglabel{Sparse Depth} & \imglabel{(a) CSPN \cite{cspn}} & \imglabel{(b) NLSPN \cite{nlspn}} & \imglabel{(c) \netname{}} \\
    \end{tabular}
    \caption{\textbf{Qualitative results on NYU-Depth v2.} CSPN and NLSPN, when processing 5 points or the shifted grid pattern, manifest the complete inability to handle them, while \netname{} maintains the scene structure.}
    \label{fig:nyu qualitative}
 \end{figure*}
 \endgroup

The bottom portion of Table \ref{tab:sota comparison nyu} reports the outcome of the evaluation with different spatial distributions and their average density of depth input points. Specifically, we report results with the two distributions depicted in the rightmost images of Figure \ref{fig:nyu-patterns}. From the table, we can observe that when the spatial distribution covers the whole image, as in the case of the Livox-like pattern, \netname{} and NLSPN achieve similar performance while other methods fall behind. However, when the input points do not cover significant portions of the scene and the density decreases further, like in the shifted-grid case, our method dramatically outperforms all competitors by a large margin.     

Figure \ref{fig:nyu qualitative} shows qualitatively how \netname{} compares to CSPN and NLSPN on an NYU sample when using 500 random points, 5 points and the shifted grid. It highlights how only our method yields meaningful and compelling results with 5 points and the shifted grid, leveraging the image content much better than competitors, thanks to the proposed architectural design. At the same time, our network achieves results comparable to competitors with 500 randomly distributed points. {This fact further highlights that the robustness of \netname{} is traded with the capacity of entirely leveraging the sparse depth information when fully available.}

 \begin{table}[t]
    \centering
    \scalebox{0.70}{
    \begin{tabular}{l | c | c c c c c}
        \hline
        Method & Lines      & RMSE (mm) $\downarrow$  & MAE $\downarrow$                               \\
        \hline
         NLSPN \cite{nlspn}              & \mr{5}{64}        & \textbf{778.00}    & \textbf{199.50}    \\
         pNCNN \cite{pncnn}              &                   & 1011.86            & 255.93             \\
         PackNet-SAN \cite{packnet-san}  &                   & 1027.32            & 356.04             \\
         PENet \cite{penet}              &                   & \underline{791.62} & 242.25             \\
         \netname{} (ours)                           &                   & 844.79             & \underline{218.39} \\
        \hline                        
         NLSPN \cite{nlspn}              & \mr{5}{32}  & \underline{1217.21}  & \underline{367.49}  \\
         pNCNN \cite{pncnn}              &             & 1766.84              & 615.93              \\
         PackNet-SAN \cite{packnet-san}  &             & 1836.84              & 914.33              \\
         PENet \cite{penet}              &             & 1853.06              & 1025.42             \\
         \netname{} (ours)                           &             & \textbf{1164.18}     & \textbf{339.22}     \\
        \hline                         
         NLSPN \cite{nlspn}              & \mr{5}{16}  & \underline{1988.52}  & \underline{693.10}  \\
         pNCNN \cite{pncnn}              &             & 3194.69              & 1321.74             \\
         PackNet-SAN \cite{packnet-san}  &             & 2841.35              & 1570.05             \\
         PENet \cite{penet}              &             & 3538.02              & 2121.46             \\
         \netname{} (ours)                           &             & \textbf{1863.25}     & \textbf{606.92}     \\
        \hline
         NLSPN \cite{nlspn}              & \mr{5}{8}   & 3234.93             & \underline{1491.28}  \\
         pNCNN \cite{pncnn}              &             & 5921.94             & 2999.92              \\
         PackNet-SAN \cite{packnet-san}  &             & \underline{3231.03} & 1575.41              \\
         PENet \cite{penet}              &             & 6015.02             & 3812.45              \\
         \netname{} (ours)                           &             & \textbf{2691.34}    & \textbf{1087.21}     \\
        \hline                         
         NLSPN \cite{nlspn}              & \mr{5}{4}    & \underline{4834.22} & 2742.80              \\
         pNCNN \cite{pncnn}              &              & 9364.58             & 5362.45              \\
         PackNet-SAN \cite{packnet-san}  &              & 4850.20             & \underline{2255.08}  \\
         PENet \cite{penet}              &              & 9318.86             & 5819.36              \\
         \netname{} (ours)                           &              & \textbf{3533.74}    & \textbf{1622.64}     \\
        \hline                         
    \end{tabular}
    }
    \caption{\textbf{Evaluation on KITTI DC.} Comparison with state-of-the-art methods, always trained on 64 lines Velodyne lidar and tested with a different number of lines. In bold is the best result, underlined the second one.}
    \label{tab:sota comparison kitti}
\end{table}

\textbf{Results on KITTI DC.} Once we assessed the performance on the indoor NYU dataset, we report in Table \ref{tab:sota comparison kitti} the evaluation on KITTI DC. 
From the table, we can notice that with 64 lines, \netname{} results almost comparable to the best one, NLSPN. However, by reducing the number of lines from 32 to 4, our network gets always the best performance with an increasing gap. Interestingly, PackNet-SAN \cite{packnet-san}, which has been specifically trained to perform well in both depth completion (64 lines) and depth prediction (0 lines) is not able to deal with fewer lines. Indeed, the accuracy it achieves when processing 16, 8 or 4 lines is even lower than the one achieved when performing depth prediction, i.e. with RMSE equal to 2.233 mm. 
We ascribe this behaviour to the fact that they train an external encoding branch to extract features from sparse data and feed them to the network by means of a sum operation. Even though such a branch applies a special and bulky sparse convolution operator \cite{MinkoswkiConvs}, it does not seem capable of generalizing to fewer points. On the contrary, the whole network seems to suffer of the same issues of fully convolutional models, resulting effective only when fed with 64 LiDAR lines or none -- the only configurations observed during training.

Figure \ref{fig:kitti qualitative} shows, on an image of the KITTI DC dataset and for three different numbers of lines, the outcome of NLSPN, PENet and our network. In contrast to competitors, \netname{} consistently infers meaningful depth maps, even when the number of lines decreases. This behaviour can be perceived better by looking at the error maps. For instance, it is particularly evident with 4 lines, focusing on the road surface and the far and background objects.   

Additional qualitative results are reported as videos in the \textbf{supplementary material} and in our project page.

\newcommand\kittiqt[3]{
    \begin{subfigure}[b]{#1\linewidth}
        \includegraphics[width=\linewidth]{imgs/kitti-qualitative/#2/#3.png}
    \end{subfigure}
}
 
\renewcommand{\tabcolsep}{2pt}
\begin{figure*}[t]
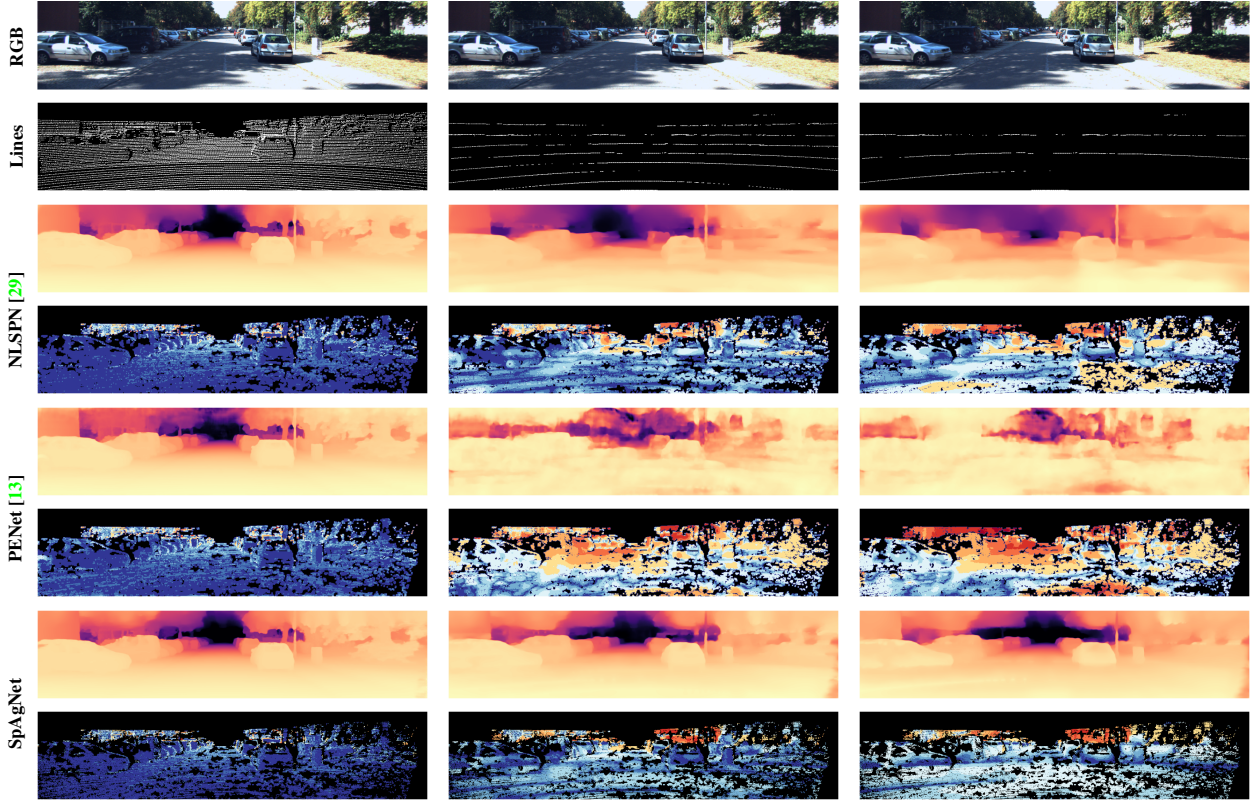

    \centering
    \begin{tabular}{ccccc}
        \rotatebox[origin=l]{90}{\scriptsize \quad \textbf{RGB}} & 
        \kittiqt{0.3}{hints}{rgb} & \kittiqt{0.3}{hints}{rgb} & \kittiqt{0.3}{hints}{rgb} & \\
        \rotatebox[origin=l]{90}{\scriptsize \quad \textbf{Lines}} &
        \kittiqt{0.3}{hints}{lines64} & \kittiqt{0.3}{hints}{lines8} & \kittiqt{0.3}{hints}{lines4} \\
        \multirow{2}{*}{\rotatebox[origin=c]{90}{\scriptsize \textbf{NLSPN \cite{nlspn}}}} &
        \kittiqt{0.3}{nlspn}{lines64}  & \kittiqt{0.3}{nlspn}{lines8}  &  \kittiqt{0.3}{nlspn}{lines4} \\
        & \kittiqt{0.3}{nlspn}{err64}  &  \kittiqt{0.3}{nlspn}{err8}   &   \kittiqt{0.3}{nlspn}{err4} \\
        \multirow{2}{*}{\rotatebox[origin=c]{90}{\scriptsize \textbf{PENet \cite{penet}}}} &
        \kittiqt{0.3}{penet}{lines64} &  \kittiqt{0.3}{penet}{lines8}  &  \kittiqt{0.3}{penet}{lines4} \\
        & \kittiqt{0.3}{penet}{err64}   &  \kittiqt{0.3}{penet}{err8}    &  \kittiqt{0.3}{penet}{err4} \\
        \multirow{2}{*}{\rotatebox[origin=c]{90}{\scriptsize \textbf{\netname{}}}} &
        \kittiqt{0.3}{our}{lines64}   &  \kittiqt{0.3}{our}{lines8}   &   \kittiqt{0.3}{our}{lines4} \\
        & \kittiqt{0.3}{our}{err64}   &  \kittiqt{0.3}{our}{err8}   &   \kittiqt{0.3}{our}{err4} \\
    \end{tabular}
    \caption{\textbf{Qualitative results on KITTI DC.} We report results using, respectively, from left to right, 64, 8 and 4 lines. From top to bottom the predicted depth and error map of \cite{nlspn}, \cite{penet} and ours.}
    \label{fig:kitti qualitative}
 \end{figure*}

\newcommand\cmark{\ding{51}}
\newcommand\xmark{{\color{red}\ding{55}}}

\begin{table}[t]
    \centering
    \scalebox{0.60}{
    \begin{tabular}{c c c}
    
    \begin{tabular}{c c c | c | c}
    \hline
    NLSP       & Confidence & Scaling    & Samples    & RMSE (m) $\downarrow$ \\
    \hline
    \xmark     & \xmark     & \xmark     &            & 0.161             \\
    \xmark     & \cmark     & \cmark     & \mr{8}{500}& 0.127             \\
    \cmark     & \xmark     & \cmark     &            & 0.122             \\
    \cmark     & \cmark     & \xmark     &            & 0.115             \\
    \cmark     & \xmark     & \xmark     &              & 0.132             \\
    \xmark     & \cmark     & \xmark     &              & 0.145             \\
    \xmark     & \xmark     & \cmark     &              & 0.135             \\
    \cmark     & \cmark     & \cmark     &            & \textbf{0.114}    \\
    \hline
    \xmark     & \xmark     & \xmark     &            & 0.770              \\
    \xmark     & \cmark     & \cmark     & \mr{8}{5}  & 0.474             \\
    \cmark     & \xmark     & \cmark     &            & 0.479             \\
    \cmark     & \cmark     & \xmark     &            & 0.526             \\
    \cmark     & \xmark     & \xmark     &             & 0.566              \\
    \xmark     & \cmark     & \xmark     &             & 0.823              \\
    \xmark     & \xmark     & \cmark     &             & 0.484              \\
    \cmark     & \cmark     & \cmark     &            & \textbf{0.467}    \\
    \hline
    \end{tabular}
    & \quad\quad &
    \begin{tabular}{l c | c | c c}
    \hline
    Backbone    & Size & Samples     & RMSE (m) $\downarrow$ \\
    \hline
    ResNet18    & 27M        & \mr{6}{500} & 0.116            \\
    ResNet34    & 37M        &             & 0.121            \\
    ResNet50    & 51M        &             & 0.117            \\
    ResNeXt50   & 51M        &             & \textbf{0.114}   \\
    DenseNet121 & 30M        &             & 0.118            \\
    DenseNet161 & 61M        &             & 0.115            \\
    \hline
    ResNet18    & 27M        & \mr{6}{5}   & 0.504            \\
    ResNet34    & 37M        &             & 0.474            \\
    ResNet50    & 51M        &             & 0.664            \\
    ResNeXt50   & 51M        &             & \textbf{0.467}   \\
    DenseNet121 & 30M        &             & 0.678            \\
    DenseNet161 & 61M        &             & 0.564            \\
    \hline 
    \multicolumn{4}{c}{}\\
    \multicolumn{4}{c}{}\\
    \multicolumn{4}{c}{}\\
    \multicolumn{4}{c}{}\\
    \end{tabular} \\
    \textbf{(a)} & & \textbf{(b)} \\
    \end{tabular}
    }
    \caption{\textbf{Ablation study on NYU -- (a) single components, (b) different backbones.} Training with 500 points, testing either with 500 or 5 points on the same dataset.}
    \label{tab:ablation study}
\end{table}

\subsection{Ablation Study}
\label{sec:ablation study}

Finally, we carry out an ablation study concerning the main components of \netname{} to measure their effectiveness. Specifically, in Table \ref{tab:ablation study}, we conduct two main studies, respectively, to evaluate (a) the impact of i) the Scale step of the S\&P modules (while the Place step is strictly necessary, being it the entry point to input the sparse depth points needed to perform completion), ii) the usage of confidence and iii) the non-local propagation head, and (b) results achieve with different backbones. 

From (a), we can notice that with 500 sparse points, scaling does not significantly improve since the network already learns to generate an output that is almost in scale. However, with only 5 points, applying a global scaling procedure helps retrieve the correct scale even in regions lacking depth measurements. 
Focusing on confidence, it turns out to be effective with high and low densities of input points. Finally, Non-Local Spatial Propagation further boosts performance in both cases.

In (b), most backbones yield comparable results when tested with 500 points, with ResNeXt50 achieving slightly better results. A significant gap in accuracy emerges when testing the same networks with  only 5 points, with ResNeXt50 achieving the best results again.

\section{Conclusion}
\label{sec:conclusions}

This paper proposes a sparsity agnostic framework for depth completion relying on a novel Scale and Place (S\&P) module. Injecting sparse depth points to it rather than to convolutions allows us to improve the robustness of the architecture even when facing uneven and sparse distributions of input depth points. In contrast, existing state-of-the-art solutions are not robust in such circumstances and are often unable to infer meaningful results. Experimental results demonstrate the ability of our network to be competitive with state-of-the-art facing standard input distributions, while resulting much better when dealing with uneven ones.

\textbf{Acknowledgement.} We gratefully acknowledge Sony Depthsensing Solutions SA/NV for funding this research and Valerio Cambareri for the constant supervision through the project and his feedback on this manuscript.

{\small
\bibliographystyle{ieee_fullname}
\bibliography{bibliography}
}
\end{document}